\documentclass{article}
\pdfoutput=1

\usepackage[nonatbib,final]{main}

\usepackage[utf8]{inputenc} 
\usepackage[T1]{fontenc}    
\usepackage{hyperref}       
\usepackage{url}            
\usepackage{booktabs}       
\usepackage{amsfonts}       
\usepackage{nicefrac}       
\usepackage{microtype}      
\usepackage{xcolor}         
\usepackage{svg} 

\usepackage{verbatim}

\usepackage{cite}
\usepackage{amsmath,amssymb}
\usepackage{algorithmic}
\usepackage{graphicx}
\usepackage{textcomp}

\usepackage{tikz}
\usepackage{bibnames}
\usepackage{multirow}
\usepackage{float} 
\usepackage{ragged2e} 
\usepackage{caption}
\usepackage{subcaption}
\usepackage{todonotes}

\usetikzlibrary{decorations.pathmorphing}

\title{Evaluating Meta-Feature Selection for the Algorithm Recommendation Problem}

\author{
Gean T. Pereira, Moisés R. dos Santos, André C. P. L. F. Carvalho\\ 
Computer Science Department, Institute of Mathematics and Computer Sciences (ICMC)\\
University of São Paulo, São Carlos, Brazil\\
{\tt\small \{geantrinpereira,mmrsantos,edesio\}@usp.br}\\
{\tt\small andre@icmc.usp.br}
}

\begin{document}

\maketitle

\begin{abstract}
With the popularity of Machine Learning (ML) solutions, algorithms and data have been released faster than the capacity of processing them. 
In this context, the problem of Algorithm Recommendation (AR) is receiving a significant deal of attention recently. This problem has been addressed in the literature as a learning task, often as a Meta-Learning problem where the aim is to recommend the best alternative for a specific dataset. For such, datasets encoded by meta-features are explored by ML algorithms that try to learn the mapping between meta-representations and the best technique to be used. One of the challenges for the successful use of ML is to define which features are the most valuable for a specific dataset since several meta-features can be used, which increases the meta-feature dimension. This paper presents an empirical analysis of Feature Selection and Feature Extraction in the meta-level for the AR problem. The present study was focused on three criteria: predictive performance, dimensionality reduction, and pipeline runtime. 
As we verified, applying Dimensionality Reduction (DR) methods did not improve predictive performances in general. However, DR solutions reduced about 80\% of the meta-features, obtaining pretty much the same performance as the original setup but with lower runtimes. 
The only exception was PCA, which presented about the same runtime as the original meta-features. Experimental results also showed that various datasets have many non-informative meta-features and that it is possible to obtain high predictive performance using around 20\% of the original meta-features. Therefore, due to their natural trend for high dimensionality, DR methods should be used for Meta-Feature Selection and Meta-Feature Extraction.
\end{abstract}

\section{Introduction}

The machine learning popularity in automated systems is increasing daily \cite{jordan2015}. The results in the rise of machine learning applications in different areas of interest that further expanded to meta-learning problems~\cite{brazdil2008metalearning}. Meta-learning is a sub-area of machine learning that investigates the problem of learning to learn~\cite{hutter2019automatic}. The aim is to develop a model that can learn new skills and rapidly adapt to a new environment, by inspecting the relationship among problems and performance of learning algorithms~\cite{finn2017model}. This association is further explored for algorithm selection, which leads to the algorithm recommendation problem, in which the most suitable algorithm is recommended according to the dataset~\cite{alcobacca2018dimensionality}. Specialists usually evaluate a group of algorithms for suitability to identify the most effective recommendation algorithm for a given case. These algorithms are trained using historical data, and their efficiency is assessed using cross-validation methods. The best algorithm is selected based on having the lowest predictive error or highest accuracy.

There has been a large amount of research on constructing meta-level problems and models~\cite{pereira2019transfer,aguiar2019meta,campos2019machine}. Nevertheless, few attention has been given to the usefulness and discriminating power of dataset characteristics to identify features that are most critical for obtaining good results~\cite{alcobacca2018dimensionality}. To found the most representative feature in a dataset, Feature Selection (FS) is often employed. In this paper, we evaluate FS methods for the Algorithm Recommendation Problem. FS is a crucial step in data mining. It is a Dimensionality Reduction (DR) technique that decreases the number of features to a convenient size for processing and analysis. FS is different from other DR techniques; it does not change the original feature set; instead, it selects a subgroup of relevant features by excluding all other features whose existence in the dataset does not positively affect the learning model~\cite{guyon2008feature}. 

To construct a model, the only source of information used for learning is the set of features. Thus, it is crucial to find the optimal subset of features that represent the original dataset. However, selecting that optimal subset is not a trivial task; if a large number of features are selected, time and space complexity increases, causing a high workload on the classifier and a decrease in performance~\cite{flach2012machine}. On the other hand, if too few features are selected, then there is a possibility of losing useful information, which again leads to a decrease in model performance. Thus, an optimal subset of relevant features is needed, which gives the best solution without decreasing the model's performance \cite{marsland2015machine}. But there is no single feature selection method that is best all the time. Specialists typically have to experiment with different methods using a trial and error approach, but the problem is that this is time-consuming and costly, especially when we have large datasets. This work aims to present an empirical analysis of feature selection and feature extraction in the meta-level setup for datasets related to the algorithm recommendation problem. The present study performs three comparison criteria: predictive performance, dimensionality reduction, and pipeline runtime.

The remaining of this work is organized as follows: Sections \ref{sec:related_works1} and \ref{sec:related_works2} shows the background related to Meta-Learning for Algorithm Recommendation and Feature Selection methods; Section \ref{sec:experimental_methodology} shows the experimental setup and methodology; The results and proper discussion are presented in Section \ref{sec:results}; Our conclusion and future work directions are discussed in Section \ref{sec:conclusion}.

\section{Meta-learning for Algorithm Recommendation}\label{sec:related_works1}

The work \cite{rice1976algorithm} was purposed the algorithm selection problem, as know as the algorithm recommendation problem.  The problem can be enunciated as the mapping of a set of data instances to a set of candidate algorithms for optimizing a specific performance metric. It seems intuitive that similar data sets might have a good solution with the same algorithm. However, it is necessary to know if exists an algorithm that has the best performance for all problem domains. This question was explored by \cite{wolpert1996lack} in the "no free lunch" theorems. For simplicity, one of the theorems demonstrates that, given a set of algorithms applied to all possible domains, the average performance of the algorithms is the same.  This theorem for supervised machine learning implies that not exists a unique learning algorithm with performance significant superior for all domains.

The best solution is to test all of the algorithm candidates for a specific task and choose to achieve the best performance. However, exists many limitations to adopt this approach. One of the main limitations is the high computational cost that can turn impossible to have a solution on available time \cite{brazdil2008metalearning}. A possible solution is to use meta-learning for the algorithm recommendation problem \cite{brazdil2008metalearning}.  A contemporary definition for meta-learning or "learn to learning" is the set of methods that consists of using the knowledge extracted from the tasks, algorithms, or model performance evaluation of past datasets to perform better, faster, and more efficient tasks for new datasets \cite{finn2017model}.  

Although many approaches exist for meta-learning model selection, in this work, we focus on meta-feature-based meta-learning\cite{hutter2019automatic}, where, given a set of datasets and algorithm candidates, the main idea is to model the algorithm recommendation problem as a supervised machine learning problem \cite{vilalta2002perspective}. The predictive attributes or meta-features are the set of characteristics extracted from a dataset, and the target is the algorithm candidate that achieved better performance for this dataset. Thus, the aim is to build a model able to recommend the best algorithm for new tasks.

Meta-learning has been successfully applied to
a large number of learning tasks and domains \cite{kanda2011using} \cite{parmezan2017metalearning} \cite{talagala2018meta} \cite{aguiar2019meta}. However, with the rising number of purpose meta-features in the literature \cite{rivolli2018towards}\cite{rivolli2018characterizing}, it is necessary to study feature selection methods to improve the meta-learning setup efficiency.

\section{Feature Selection Methods} \label{sec:related_works2}

It is common sense that features used for training models have a significant influence on the performance. With that said, irrelevant features can harm performance. Thus, feature selection methods are employed to found the most relevant features in a training dataset, which contribute more to the target variable. Using feature selection can lead to less redundant data, which lower the probability of making a decision based on noisy feature, so could reduce overfitting and also improve the model's accuracy. Dimensions of the training set also decrease because feature selection chooses a relevant subset of features within a whole dataset; thus, it also reduces the time complexity and algorithms train faster. There are two types of feature selection methods, model-based feature selection, and filter-based filter selection. This work focuses on filter-based feature selection because these are methods with a low computational cost. In the next sections, some methods for filter-based feature selection are shortly discussed. For more information read \cite{lazar2012survey} and \cite{bommert2020benchmark}.

\subsection{Variance Threshold}
This feature selection method removes features with a low variance; thus, it calculates each feature's variance, then drops features with variance below some threshold, making sure that the features have the same scale. Although it is a simple method, it has a compelling motivation, since the idea is that low variance features contain less relevant information. Moreover, the variance threshold is an unsupervised method, since it only looks at the feature values and not to the desired output. 

\subsection{Univariate Analysis}
This feature selection method consists of selecting the best features based on statistical tests. It is often used as a pre-processing step to an estimator, thus improving the performance of a model and cutting of the time complexity, since it reduces the dimensions of a dataset. These statistical tests are used to found the relationship between each input feature and the output feature. Those input feature vectors are kept and used for further analysis, which gives a strong statistical relationship with the output variable, and remaining input features, whose relationship is weak, are discarded. There is much statistical test used in the univariate feature selection method i.e., Chi-squared, Pearson correlation, and ANOVA F-value test, and others. Two of them are described below: 

\begin{itemize}
    \item \textbf{Chi-squared test:} A statistical test that aims to evaluate the dependency between two variables. Although it has some common similarities with the coefficient of determination, the chi-square test is only valid for nominal or categorical data, whereas the determination coefficient is only valid for numeric data. This method has various applications, and one of them is feature selection. Considering an input feature variable and an output variable, i.e., class label, first, the Chi-squared statistics of every input feature variable concerning the output is calculated to get the dependency relationship between them. Afterward, features that have shown less dependency with output variables are discarded, while the features that have shown strong dependency with the target variable are kept to training the model. 
    
 \item \textbf{ANOVA F-test:} This method is based on F-test, which is used only for quantitative data. ANOVA F-test estimates the degree of linear dependency between an input variable and the target variable, giving a high score to the features which are highly correlated and low score to less correlated features. The F-value scores inspect if the means for each group are considerably different when we group the numerical feature by the target vector.
\end{itemize}

\section{Experimental Methodology}\label{sec:experimental_methodology}
The datasets used in this work were collected from the Aslib repository~\cite{bischl2016aslib}, being 28 meta-datasets related to the algorithm recommendation problem. 
These meta-datasets are related to classic optimization problems, such as the Traveling Salesman
Problem (TSP) and the Propositional Satisfiability Problem (SAT). Table \ref{tab:data} shows the dataset names and basic summary of them.

\begin{table}[H]
\begin{minipage}{.49\linewidth}
    \centering
    \caption{Dataset's statistics.}
    \resizebox{0.9\textwidth}{!}{
    \begin{tabular}{l c c c}
    \toprule
         \textbf{Dataset} & \textbf{Instances} & \textbf{Features} & \textbf{Classes} \\
    \toprule
         ASP-POTASSCO     & 1294               & 139               & 11               \\  
         BNSL-2016        & 1179               & 93                & 8                \\  
         CPMP-2015        & 527                & 22                & 4                \\  
         CSP-2010         & 2024               & 67                & 2                \\  
  CSP-Minizinc-Obj-2016   & 100                & 123               & 3                \\
  CSP-Minizinc-Time-2016  & 100                & 117               & 4                \\
         CSP-MZN-2013     & 4636               & 117               & 10               \\
         GRAPHS-2015      & 5723               & 36                & 6                \\
         MAXSAT-PMS-2016  & 596                & 44                & 12                \\
         MAXSAT-WPMS-2016 & 630                & 53                & 10                \\
         MAXSAT12-PMS     & 876                & 31                & 6                \\
      MAXSAT15-PMS-INDU   & 601                & 57                & 16                \\
         MIP-2016         & 214                & 120               & 3                \\
         OPENML-WEKA-2017 & 105                & 157               & 3                \\
         PROTEUS-2014     & 4021               & 32                & 22                \\
         QBF-2011         & 1368               & 46                & 5                \\
         QBF-2014         & 1248               & 46                & 13                \\
         QBF-2016         & 825                & 67                & 14                \\
         SAT03-16\_INDU    & 2000               & 139               & 10                \\
         SAT11-HAND       & 296                & 66                & 9                \\
         SAT11-INDU       & 300                & 67                & 11               \\
         SAT11-RAND       & 592                & 53                & 8                \\
         SAT12-ALL        & 1614               & 58                & 29                \\
         SAT12-HAND       & 767                & 70                & 22                \\
         SAT12-INDU       & 1167               & 74                & 21                \\
         SAT12-RAND       & 1362               & 98                & 9                \\
         SAT15-INDU       & 300                & 87                & 10                \\
         TTP-2016         & 9714               & 58                & 18                \\
    \bottomrule
    \label{tab:data}
    \end{tabular}
    }
\end{minipage} 
\enskip
\begin{minipage}{.49\linewidth}
\centering
\caption{Classifiers and their hyperparameters.}
\resizebox{0.9\textwidth}{!}{
\begin{tabular}{l l c c}
\toprule
\textbf{Classifier}  & \textbf{Hyperparameter} & \textbf{Default} & \textbf{Range}                \\
\toprule
\rule{0pt}{1ex}
KNN                  & Num nearest neighboors  & 3                & (1, 51)                             \\

\rule{0pt}{4ex}
\multirow{3}{*}{DT}  & Min samples split       & 2                & (2, 51)                       \\
                     & Min samples leaf        & 1                & (2, 51)                       \\
                     & Max depth               & None             & (2, 31)                       \\

\rule{0pt}{4ex}
\multirow{2}{*}{SVM} & C                       & 1                & (1, 32769)                    \\
                     & Gamma                   & 1/Nº features     & (1, 32769)                    \\

\rule{0pt}{4ex}
\multirow{2}{*}{RF}  & Nº estimators           & 10               & (1, 1025)                     \\
                     & Max depth               & None             & (1, 21)                       \\
    \bottomrule
    \label{tab:classifiersHyperparameters}
    \end{tabular}
    }
\end{minipage} 
\end{table}


Feature Selection methods used in this work were implemented using ``scikit-learn''~\cite{scikit-learn}. For Variance Threshold, features that do not change in more than $80$, $85$, $90$, and $95$ percent of the observations are removed. For Chi-squared and ANOVA, features are select according to a percentile with the highest score for each feature selection method. The percentiles considered were $80$, $85$, $90$, and $95$.

For comparison criterion, the feature extraction technique "Principal Component Analysis" (PCA) adds to this experiment. This technique achieved the best results in the work of \cite{alcobacca2018dimensionality}. The selected components correspond to that maintain 80, 85,90, and 95 of the original observations. These values are stemming from \cite{alcobacca2018dimensionality}, and the PCA implementation is available on the "scikit-learn" package~\cite{scikit-learn}.


Table~\ref{tab:classifiersHyperparameters} shows the classifiers used (\textbf{Classifier}), such as K-Nearest Neighbors (KNN), Decision Tree (DT), Support Vector Machine (SVM), and Random Forest (RF), their hyperparameters tuned (\textbf{Hyperparameter}), the default value for the specific hyperparameter tuned (\textbf{Default}) and the range values tested during the tuning of that hyperparameter (\textbf{Range}). Also, in Table~\ref{tab:classifiersHyperparameters}, values ``None'' in DT and RF rows are the default values defined by Scikit-learn, which are said to be expanded until all leaves are pure or until all leaves contain fewer samples than \textit{Min samples split} parameter. 


For model evaluation, all the meta-datasets were scaled between 0 and 1 by Minimum-Maximum scaler \cite{han2011data}. The pipeline includes feature selection methods, meta-learners, and all parameters for the hyperparameter tuning process.  
Hyperparameter tuning is made by the Random Search K-Fold cross-validation technique in which we split our data into $k$ subsets, and then we tune and train our models on $k-1$ folds, and the performance metric are apply on the left fold. This process is repeated k times until all the folds have been evaluated. Given a hyperparameter space of a specific machine learning model, cross-validation is applied $n$ times with a random variation of hyperparameters space. Thus, the hyperparameters set with the best performance is selected to be applied to test data \cite{bergstra2012random}. The number of $k$ subsets used were $10$, a value widely used on the literature \cite{russell2009inteligencia}.

The models are evaluated in three criteria: predictive performance, proportion of dimensionality reduction, and time to fit. For performance, uses the balanced accuracy metric \cite{brodersen2010balanced}. Balanced accuracy is a fair version of the accuracy metric for multi-class problems. The proportion of dimensionality reduction is the difference between the number of dimensions of the original meta-dataset and the dimensions after feature selection or extraction, divided by the number of dimensions of the original meta-dataset. Time to fit is the average time to induce the feature selection or extraction and the classification model. This time is returned by the random search cross-validation function of the ``scikit-learn'' package. For reproducibility, all the paper experiment's implementation is public on this repository: \url{https://github.com/moisesrsantos/featselec}. 

\section{Results and Discussion}\label{sec:results}
This section discuss the results of applying 4 meta-learners (SVM, KNN, Decision Tree, and Random Forest) to 28 meta-datasets related to algorithm recommendation, which were also submitted to 3 Feature Selection methods (Chi-squared, F-ANOVA and Variance threshold) and 1 Feature Extraction method (PCA). 
Our experiments were performed to answer three research questions: 
(1) Is there any predictive performance improvement of applying feature selection to meta-datasets related to algorithm recommendation? 
(2) How good is feature selection to reduce the dimensionality of those meta-datasets? 
(3) Is there any impact on the pipeline execution time when using feature selection?     

The following results are divided in three parts, each one aiming to answer one of the research questions. 
We used bar plots and box-plots to discuss the properties of the results distributions, as well as Critical Difference (CD) diagrams to better show the statistical results between comparisons. 
In Figures~\ref{fig:nemenyi_performance}, \ref{fig:nemenyi_reduction}, and \ref{fig:nemenyi_time}, the diagrams illustrate the Friedman-Nemenyi test with 95\% of confidence for the Balanced Accuracy, percentage of dimensionality reduction, and pipeline execution times. 
In the CD diagrams, the more towards to 1, the better is the method. 
Therefore, the methods on the left have better performances than the methods on the right. 
Besides, it is important to consider the CD intervals that cross the lines representing the methods.
These intervals demonstrate that all the lines that touch it have statistically equal values, so no method is statistically better than the others and, thus, they all have comparable performances.    


%
%
\subsection{Predictive Performance}
\begin{figure}[htbp]
\begin{minipage}{.47\linewidth}
\centering
\includegraphics[width=0.95\textwidth]{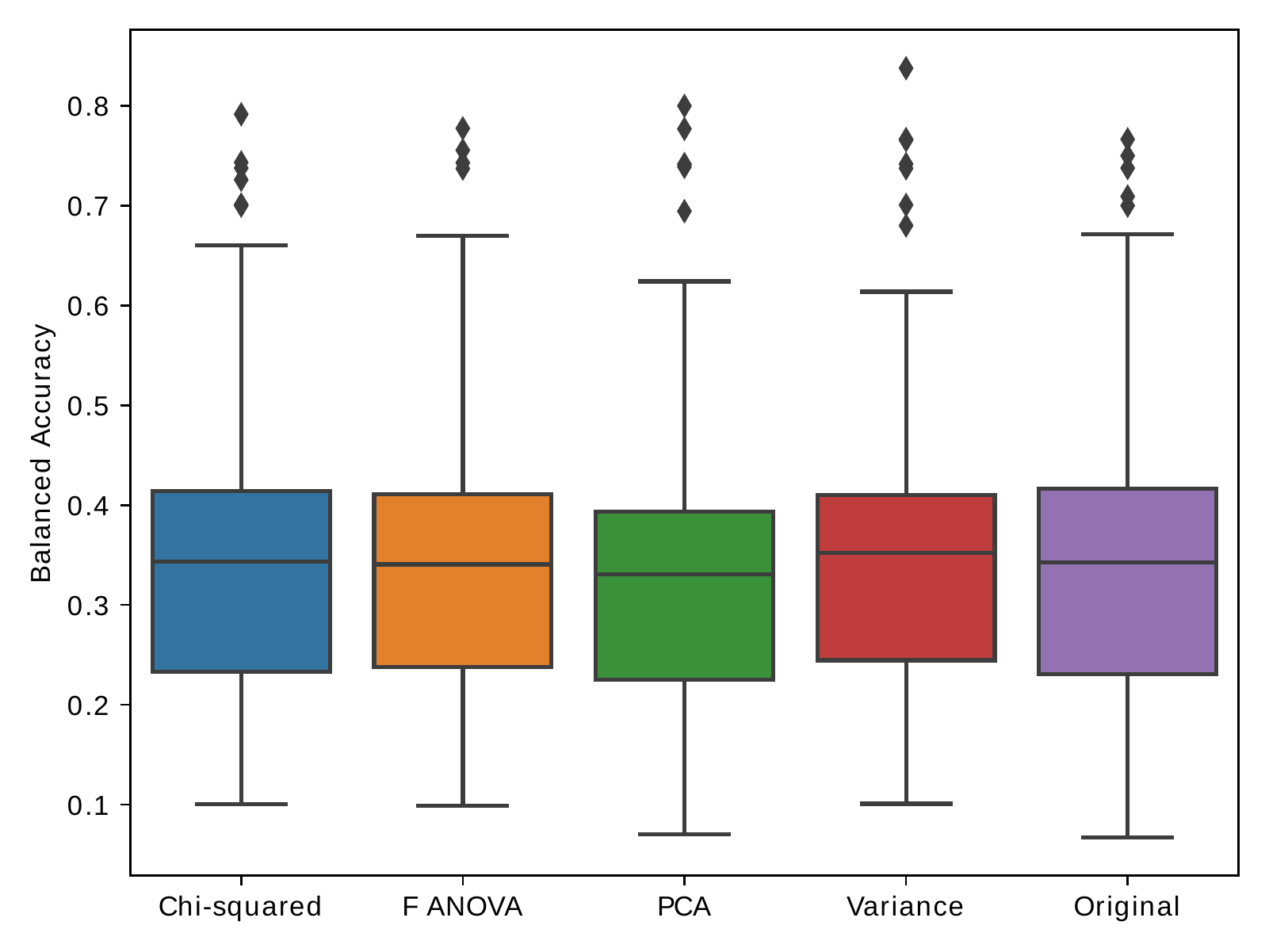}
\caption{Balanced Accuracy comparison between dimensionality reduction methods.} 
\label{fig:boxplot_performances}
\end{minipage} 
\enskip
\begin{minipage}{.51\linewidth}
\centering
\resizebox{\textwidth}{!}{
\begin{tikzpicture}[xscale=1.75]
\node (Label) at (1.545837121034909, 0.7){\tiny{CD = 0.58}}; 
\draw[decorate,decoration={snake,amplitude=.4mm,segment length=1.5mm,post length=0mm},very thick, color = black] (1.2,0.5) -- (1.891674242069818,0.5);
\foreach \x in {1.2, 1.891674242069818} \draw[thick,color = black] (\x, 0.4) -- (\x, 0.6);
\draw[gray, thick](1.2,0) -- (6.0,0);
\foreach \x in {1.2,2.4,3.6,4.8,6.0} \draw (\x cm,1.5pt) -- (\x cm, -1.5pt);
\node (Label) at (1.2,0.2){\tiny{1}};
\node (Label) at (2.4,0.2){\tiny{2}};
\node (Label) at (3.6,0.2){\tiny{3}};
\node (Label) at (4.8,0.2){\tiny{4}};
\node (Label) at (6.0,0.2){\tiny{5}};
\draw[decorate,decoration={snake,amplitude=.4mm,segment length=1.5mm,post length=0mm},very thick, color = black](3.3839285714285716,-0.25) -- (3.7464285714285714,-0.25);
\node (Point) at (3.4339285714285714, 0){};\node (Label) at (0.5,-0.45){\scriptsize{Original}}; \draw (Point) |- (Label);
\node (Point) at (3.519642857142857, 0){};\node (Label) at (0.5,-0.75){\scriptsize{F ANOVA}}; \draw (Point) |- (Label);
\node (Point) at (3.6964285714285716, 0){};\node (Label) at (6.5,-0.45){\scriptsize{Chi-squared}}; \draw (Point) |- (Label);
\node (Point) at (3.685714285714286, 0){};\node (Label) at (6.5,-0.75){\scriptsize{Variance}}; \draw (Point) |- (Label);
\node (Point) at (3.6642857142857137, 0){};\node (Label) at (6.5,-1.05){\scriptsize{PCA}}; \draw (Point) |- (Label);
\end{tikzpicture}
}
\caption{Critical Difference diagram of the performances.}
\label{fig:nemenyi_performance}
\end{minipage} 
\end{figure}

In order to verify whether the DR methods bring some improvement in terms of predictive performance, we analyzed the distributions of balanced accuracy obtained by applying each DR method to each meta-dataset and training with each meta-learner.  
As shown by box-plots in Figure~\ref{fig:boxplot_performances}, all DR methods had similar performances, even when compared to the original datasets.
Slightly standing out from the others, the Variance threshold appeared to reduce the variance of the predictive performance across datasets and classifiers, showing a lower range of values.
However, it also generated more sparse maximum outliers.  
Speaking of outliers, another interesting thing to observe is that the number of outliers remains pretty much the same regardless of the DR method applied, and those outliers remain located in the region of the maximum values only.

Even though Variance showed more stable results (except for its outliers), when submitted to a statistical test, it was verified that F-ANOVA had the best performance between the DR methods.
However, considering the CD interval shown in Figure~\ref{fig:nemenyi_performance}, no DR method improved the overall predictive performance in the task of algorithm recommendation with statistical significance. 
Even so, using DR was as good as using the original data.

%
%
\subsection{Dimensionality Reduction}
\begin{figure}[htbp]
\begin{minipage}{.47\linewidth}
\centering
\includegraphics[width=0.95\textwidth]{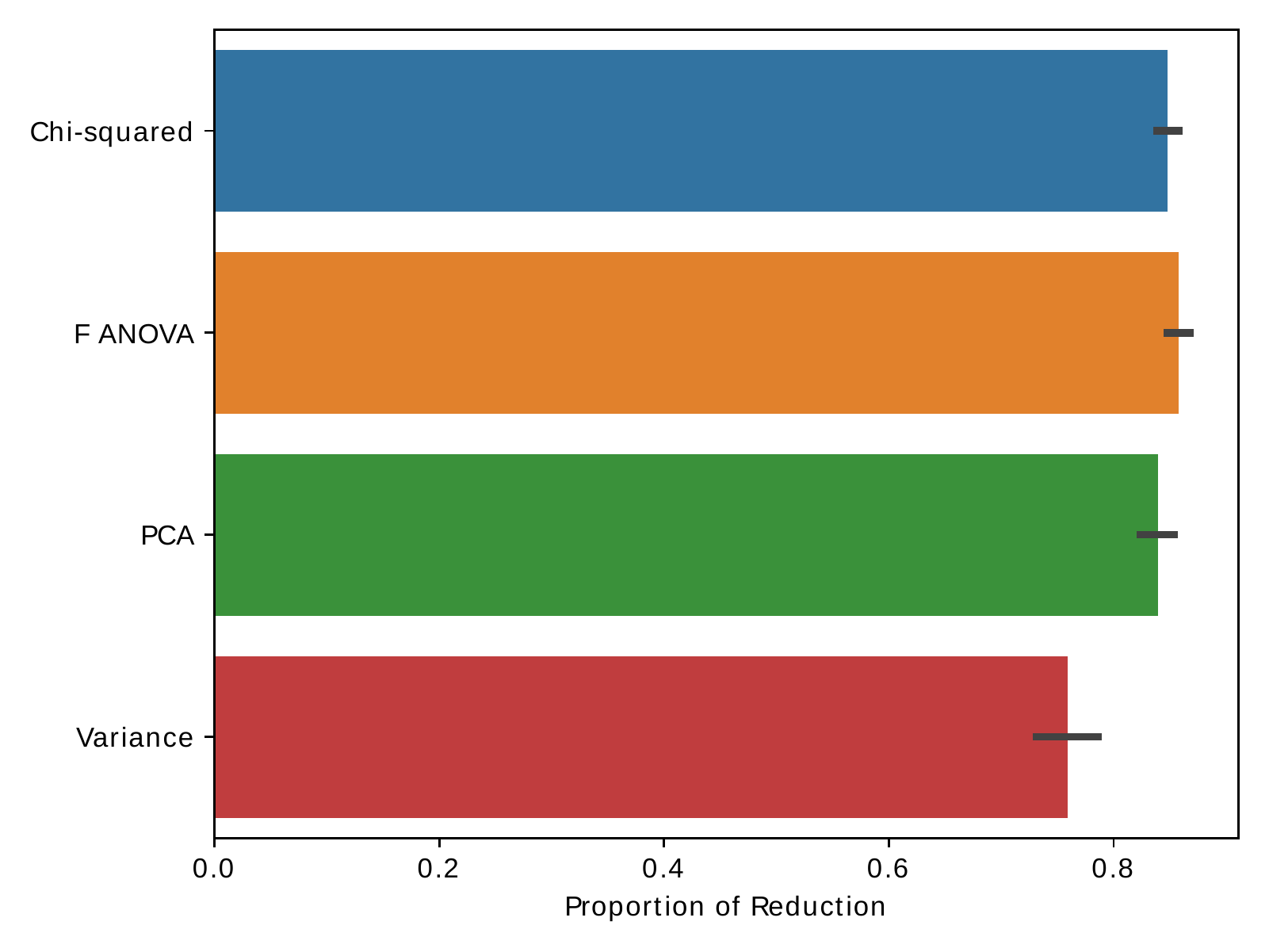}
\caption{Percentage of reduction.} 
\label{fig:barplot_reduction}
\end{minipage} 
\enskip
\begin{minipage}{.51\linewidth}
\centering
\resizebox{\textwidth}{!}{
\begin{tikzpicture}[xscale=1.75]
\node (Label) at (1.8323959541157202, 0.7){\tiny{CD = 0.44}}; 
\draw[decorate,decoration={snake,amplitude=.4mm,segment length=1.5mm,post length=0mm},very thick, color = black] (1.5,0.5) -- (2.1647919082314404,0.5);
\foreach \x in {1.5, 2.1647919082314404} \draw[thick,color = black] (\x, 0.4) -- (\x, 0.6);
\draw[gray, thick](1.5,0) -- (6.0,0);
\foreach \x in {1.5,3.0,4.5,6.0} \draw (\x cm,1.5pt) -- (\x cm, -1.5pt);
\node (Label) at (1.5,0.2){\tiny{1}};
\node (Label) at (3.0,0.2){\tiny{2}};
\node (Label) at (4.5,0.2){\tiny{3}};
\node (Label) at (6.0,0.2){\tiny{4}};
\draw[decorate,decoration={snake,amplitude=.4mm,segment length=1.5mm,post length=0mm},very thick, color = black](3.224553571428572,-0.25) -- (3.7196428571428575,-0.25);
\node (Point) at (3.2745535714285716, 0){};\node (Label) at (0.5,-0.45){\scriptsize{F ANOVA}}; \draw (Point) |- (Label);
\node (Point) at (3.5825892857142856, 0){};\node (Label) at (0.5,-0.75){\scriptsize{PCA}}; \draw (Point) |- (Label);
\node (Point) at (4.473214285714286, 0){};\node (Label) at (6.5,-0.45){\scriptsize{Variance}}; \draw (Point) |- (Label);
\node (Point) at (3.6696428571428577, 0){};\node (Label) at (6.5,-0.75){\scriptsize{Chi-squared}}; \draw (Point) |- (Label);
\end{tikzpicture}}
\caption{Critical Difference diagram of the dimensionality reduction.}
\label{fig:nemenyi_reduction}
\end{minipage}
\end{figure}

For the purpose of verifying how good are the feature selection methods in reducing the dimensionality of the selected meta-datasets, we focused on observing the performance of each technique in a broader context. Therefore, reductions in all pipelines involving each meta-dataset and all meta-learners were considered.
As seen in Figure~\ref{fig:barplot_reduction}, DR methods showed a high percentage of reduction in average, with all of them (except for Variance) achieving more than 80\% percent reduction.
Chi-squared, F-ANOVA, and PCA presented similar reduction averages in a stable manner.
In turn, Variance proved to be inferior in average and more unstable, showing higher standard deviation values. 

When considering the statistical comparisons seen in Figure~\ref{fig:nemenyi_reduction}, it is possible to notice that F-ANOVA, PCA, and Chi-squared did not differ statistically, being equally good in reducing the dimensions of the original meta-data. Furthermore, it is clear that Variance is statistically worse than others.

%
%
\subsection{Pipeline Runtime}
\begin{figure}[htbp]
\begin{minipage}{.47\linewidth}
\centering
\includegraphics[width=0.95\textwidth]{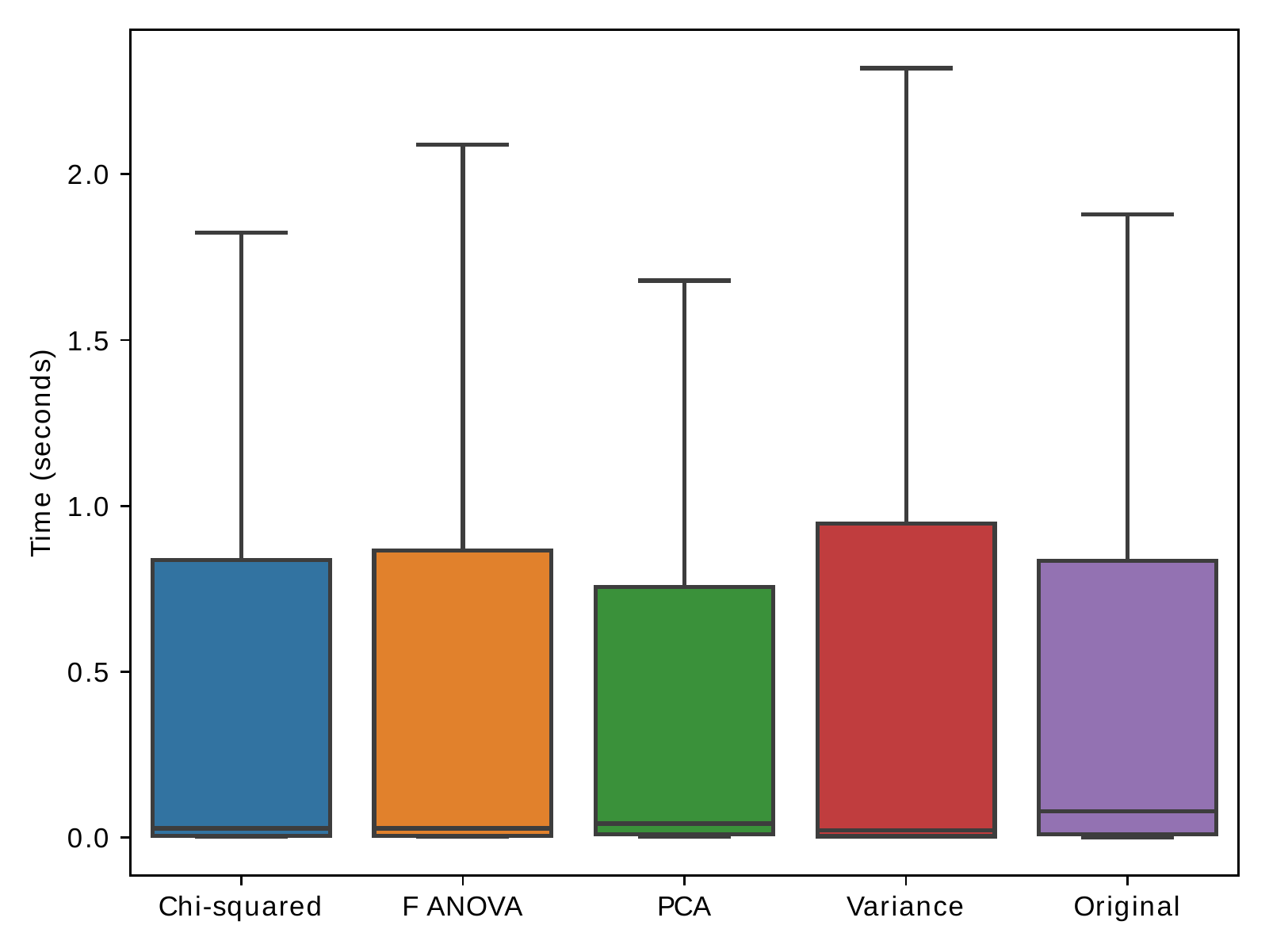}
\caption{Pipeline runtime comparison.} 
\label{fig:boxplot_time}
\end{minipage} 
\enskip
\begin{minipage}{.51\linewidth}
\centering
\resizebox{\textwidth}{!}{
\begin{tikzpicture}[xscale=1.7]
\node (Label) at (1.545837121034909, 0.7){\tiny{CD = 0.58}}; 
\draw[decorate,decoration={snake,amplitude=.4mm,segment length=1.5mm,post length=0mm},very thick, color = black] (1.2,0.5) -- (1.891674242069818,0.5);
\foreach \x in {1.2, 1.891674242069818} \draw[thick,color = black] (\x, 0.4) -- (\x, 0.6);
\draw[gray, thick](1.2,0) -- (6.0,0);
\foreach \x in {1.2,2.4,3.6,4.8,6.0} \draw (\x cm,1.5pt) -- (\x cm, -1.5pt);
\node (Label) at (1.2,0.2){\tiny{1}};
\node (Label) at (2.4,0.2){\tiny{2}};
\node (Label) at (3.6,0.2){\tiny{3}};
\node (Label) at (4.8,0.2){\tiny{4}};
\node (Label) at (6.0,0.2){\tiny{5}};
\draw[decorate,decoration={snake,amplitude=.4mm,segment length=1.5mm,post length=0mm},very thick, color = black](2.789285714285714,-0.25) -- (3.392857142857143,-0.25);
\draw[decorate,decoration={snake,amplitude=.4mm,segment length=1.5mm,post length=0mm},very thick, color = black](4.257142857142857,-0.4) -- (4.421428571428572,-0.4);
\node (Point) at (2.839285714285714, 0){};\node (Label) at (0.5,-0.65){\scriptsize{Variance}}; \draw (Point) |- (Label);
\node (Point) at (3.139285714285714, 0){};\node (Label) at (0.5,-0.95){\scriptsize{Chi-squared}}; \draw (Point) |- (Label);
\node (Point) at (4.371428571428572, 0){};\node (Label) at (6.5,-0.65){\scriptsize{PCA}}; \draw (Point) |- (Label);
\node (Point) at (4.307142857142857, 0){};\node (Label) at (6.5,-0.95){\scriptsize{Original}}; \draw (Point) |- (Label);
\node (Point) at (3.342857142857143, 0){};\node (Label) at (6.5,-1.25){\scriptsize{F ANOVA}}; \draw (Point) |- (Label);
\end{tikzpicture}}
\caption{Critical Difference diagram regarding runtimes.}
\label{fig:nemenyi_time}
\end{minipage} 
\end{figure}

To verify if there is any impact on the pipeline runtimes when using feature selection, we collected runtimes from all the pipelines performed in our experiments. Those runtimes included both the time to extract the sub-set of meta-features (feature selection) or transform the original space of meta-features (feature extraction) and the runtime to fit the data. The distributions of the runtimes for each DR method and the original setup are shown in Figure~\ref{fig:boxplot_time}. 
As seen, the times of the pipelines were short in general, with maximum values achieving close to 2 seconds.
The medians for these distributions were similar and much closer to the Q1 and the minimum. 
It is interesting to see that the variations happened mainly in the region of the maximum values, while the minimum values, Q1, and Interquartile range were similar. 

In summary, these results show that the pipelines were cheap, and the application of DR methods does not seem to have much impact on time.
However, when considering the CD diagram shown in Figure~\ref{fig:nemenyi_time}, it is possible to see that the execution times for Variance, Chi-square, and F-ANOVA, are statistically comparable and shorter than using the original data and PCA.
Using Feature Selection proved to be a better solution for reducing the overall time of pipelines than not using it, showing that, despite the additional time to lower dimensions, the total time, including the training, is shorter. On the other hand, using or not using PCA did not statistically change the runtime of the pipelines.
%
%
%
\section{Conclusion and Future Works}\label{sec:conclusion}
We presented a comparative study of Feature Selection and Feature Extraction in the meta-level setup for datasets related to Algorithm Recommendation. Our goal was to evaluate popular DR methods according to three criteria: predictive performance, dimensionality reduction, and pipeline runtime. As we verified, applying DR did not improve predictive performance in general. However, it reduced around 80\% of the meta-features yet achieving the same performance as the original setup with a shorter runtime. The only exception was PCA, which showed about the same runtime. Our experiments showed that these meta-datasets have many non-informative meta-features and that it is possible to achieve good performance using about 20\% of the original features. Traditional Meta-Learning, the one focused on hand-craft extraction of meta-features, seems to suffer a lot with the curse of dimensionality.  Instead of helping Machine Learning algorithms to achieve better or fast performance while taking advantage of high-level information (meta-level), it seems that, in some cases, meta-info hinders the task. We conclude that using DR methods is beneficial to many meta-datasets since their natural trend for high dimensionality. However, we acknowledge that more experiments with different numbers and types of meta-features are needed to ensure our claim.


\bibliographystyle{unsrt}

\end{document}